\documentclass[aip, pre]{revtex4-1}

\usepackage{amssymb}
\usepackage{amsmath}
\usepackage{graphicx}
\usepackage{float}
\usepackage{booktabs}

\begin{document}

\title{Complexity-entropy analysis at different levels of organisation in written language} 

\author{E. \surname{Estevez-Rams}}
\email{estevez@fisica.uh.cu}
\affiliation{Facultad de F\'isica-Instituto de Ciencias y Tecnolog\'ia de Materiales(IMRE), University of Havana, San Lazaro y L. CP 10400. La Habana. Cuba.}

\author{A. \surname{Mesa Rodriguez}}
\affiliation{Facultad de Matematica y Computacion, University of Havana, San Lazaro y L. CP 10400. La Habana. Cuba.}

\author{D.  \surname{Estevez-Moya}}
\affiliation{Facultad de F\'isica, University of Havana, San Lazaro y L. CP 10400. La Habana. Cuba.}

\begin{abstract}
Written language is complex. A written text can be considered an attempt to convey a meaningful message which ends up being constrained by language rules, context dependence and highly redundant in its use of resources. Despite all these constraints, unpredictability is an essential element of natural language. Here we present the use of entropic measures to assert the balance between predictability and surprise in written text. In short, it is possible to measure innovation and context preservation in a document. It is shown that this can also be done at the different levels of organization of a text. The type of analysis presented is reasonably general, and can also be used to analyze the same balance in other complex messages such as DNA, where a hierarchy of organizational levels are known to exist.
\end{abstract}

\date{\today}
\maketitle

\section{Introduction}

Human language can be argued to be the most complex of consciousness mediated systems, resulting from a long evolution process in strong and extremely involved interaction with both natural and human-constructed environment \cite{nowak01,nowak02}. When comparing to other complex systems as, for example, in non-linear dynamics, in language, there is also a balance between predictable and unpredictable behavior \cite{montemurro15}. The predictable behavior comes from those rules (e.g., grammar and semantic) that has come into place, as a common framework, that allows a given community to communicate among them. It is known that a subset of these rules can even be shared across different languages and languages families \cite{chomski65,hawkins83}. Unpredictability is associated with the human capacity to choose in interaction with its environment and accounts for what one can consider, in a broad sense, the creativity of the agent that has created the message.   However, predictability-unpredictability, as explained above, is not the only balance. There is also the question of, beyond orthographic, syntax and grammar rules, how human beings balance constraints and choice at the level of sentences and their organization, further on, at the level of the paragraph, paragraph composition and even the whole text organization as books \cite{hawkins04,debowski11,altmann11}. It is clear that at this other levels less rigid constraints are found, but yet, some rules are followed. After all, it can be agreed that texts convey meaningful messages in a highly redundant and logical way, that strives to be consistent. As a result of these different balances, an organizational hierarchy arises in language that can be readily followed in written text \cite{lacalle06,montemurro10}.

Statistics, as well as information theory, have been used to explore the information content and organization for written language \cite{shannon51,amancio13,amancio15b}. If a text is considered as a string of characters, the hierarchy organization is seen through different levels of correlation, which can be exposed by identifying different units of information: characters, words, sentences. Shannon \cite{shannon51,ebeling95} observed that not all characters convey the same information \cite{shannon48} in the sense of entropy, that is, the measure of surprise of the source that is emitting the message.  The estimate of entropy per letter in the English language is considered to be around $1$ to $1.5$ bit per character. More recent estimates rises such entropy to $2$-$2.4$ bit per character or even larger \cite{schurmann99,montemurro15}. Still, at the character level, Sch\"urmann and Grassberger \cite{schurmann07} discovered a significant dependence of character predictability on its context. They found that letter predictability depends on its position in a word, with much higher entropy at the beginning of a word and long decaying tails as further positions are considered. At the word level, the most famous finding is the one referring to the distribution of word frequency as a function of word rank which is known as Zipf(-Mandelbrot) law \cite{zipf65,mandelbrot54}. The power law dependence is taken as mathematical evidence of the complex behavior of human language. Other empirical regularities are the power law of the number of words and the length of a text, which comes under the name of Herdan-Heap law \cite{herdan64,heaps78}. 

In order to grasp the elusive idea of complexity, it is useful to consider a system, as the one described by written text, as a process that produces, stores and transmits information, much in the same spirit as has been done in other fields \cite{crutchfield89,crutchfield12}. That is the approach we follow in this paper. As mentioned before, we consider a written text as a coherent, consistent attempt to convey a rational and logical (alas understandable) message to an intelligent agent. As a result of this last statement, one expects the production of information as an essential characteristic for a human text. Storage of information also comes naturally as that contained in any block of characters of the text, while transmission of information will be understood as the correlation between different parts of a document. Both storage and transmission are needed by the reader to follow the rational and logical flow of the message from any point of the text onwards. Our analysis is aimed at measuring the balance between information production or innovation and context preservation in written language. 
 
When analyzing texts of individual authors, it may also be asked to what extent style determines the amount of information (in Shannon sense) of the texts with respect to the content being conveyed.  The question is relevant for many reasons that go from pure linguistic concerns to a broader scope beyond natural language analysis. When analyzing DNA as a code sequence, it is accepted that also a complex hierarchy of levels are found in its organizational structure. In the case of DNA, going from codons to genes and beyond, one of the essential questions, in analogy to human language, is how to identify all the semantic units and their correlation \cite{jimenez16}.  Is there a universal ``style'' in DNA writing or are there different flavors across species or genres? 

In this paper, we address the question of how to measure complexity in written text as the interrelation between disorder and structure. We will show how complexity-entropy maps can be used to assert the information content at the level of sentence, word and letter organization, allowing to separate information contribution at different levels of organization in complex symbolic sequences. We will be discussing as examples, the work of three different English language writers: W. Shakespeare, J. Abbott, and A. C. Doyle. The authors were chosen due to their different literary style. We notice that the authors shown are only to illustrate the analysis developed; similar results were found for other authors. Furthermore, the method can be generalized to other systems, including the biological world.

\section{Entropic measures of disorder and structure}

 Information theory, as laid down by Shannon \cite{shannon48}, is concerned with the degree of unpredictability of a source of random events characterized by their probabilities. Events can be considered as a random variable $X$ which takes values from a set with probabilities $\{p_i\}$. Within this framework, a measure of surprise of a given event is the negative logarithm of the probability $p_i$ of the event: $-\log_2 (p_i)$ (from now on the base $2$ of the logarithm will be dropped, and the units of information will be given by bit). The measure of the source unpredictability is then given by the expectation value of every possible event $H[X]=-\sum_i p_i \log p_i$. Also, the mutual information $I[X:Y]$  between two random variables $X$ and $Y$, or of two sources, is a  measures of the reduction of uncertainty of $X$ as result of knowing the value of $Y$ and vice-versa: $I(X:Y)=H[X]-H[X|Y]$. The second term is the unpredictability of $X$ given that $Y$ is known: $H[X|Y]=-\sum_{x,y}p(x,y)\log p(x|y)$, where $p(x,y)$ is the probability of simultaneous occurrence of event $X=x$ and $Y=y$, while $p(x|y)$ is the conditional probability that $X=x$ given that it is known that $Y=y$. If knowing the value $Y$ tell us nothing about the value of $X$, then $H[X|Y]=H[X]$ and the mutual information vanishes. If, on the other extreme, knowing the value of $Y$ determines with complete certainty the value of $X$, then $I[X:Y]=H[X]$, the largest possible value. Mutual information can also be written as $I[X:Y]=H[X]+H[Y]-H[X,Y]$, which makes explicit the symmetric character of the mutual information $I[X:Y]=I[Y:X]$  \cite{cover06}.
 
 Let us consider a written text as the result of an ergodic source that outputs symbols with a given probability distribution. Symbol output is dependent on previous symbols. As a process, we will say that the text is a particular realization of the source output given as a time series, where time is a discrete variable that index each position in the text.  Any text can then be considered a sequence of values $\ldots, s_{-1}, s_0, s_1, \ldots$, where each value belongs to the same alphabet $s_t \in \mathcal{A}$ (e.g. the Latin alphabet). In the theoretical treatment, a text is taken as bi-infinite strings: It has no starting or ending position. If the source is characterized by some probability measure, then for each block of characters of length $n$, $S^n=S_j, S_{j+1}, S_{j+n-1}$,  a probability can be associated to each particular realization $s^n=s_j, s_{j+1}, s_{j+n-1}$ of the block. For a stationary process, the probability does not depend on the position of the block in the text. The (Shannon) block entropy is then given as $H[n]=-\sum_{s^n}p(s^n)\log p(s^n)$, where the sum is taken over all possible sequences of length $n$. The entropy rate, or entropy density, $h_\mu$ is defined as
\begin{equation}
\begin{array}{ll}
 h_\mu&=\lim\limits_{n\rightarrow \infty} \frac{H[n]}{n},\label{eq:hmu}
 \end{array}
\end{equation}
which can be interpreted as the irreducible disorder of a sequence, after an infinite number of symbols have been observed, or equivalently, a measure of how unpredictable an infinite sequence looks, after considering all correlations that occur at every length scale. The entropy density is closely related to the maximum compression of the string that can be achieved by an optimum coding \cite{lz77}, the larger $h_\mu$, the closer to one of the compression ratio. It is remarkable that the entropy density $h_\mu$ for a stationary process is equal to the Kolmogorov entropy rate, the last being the length of the shortest algorithm able to reproduce the data \cite{cover06}. Entropy rate can also be understood as a measure of the rate at which the process produces new information, as any new pattern will come as a surprise for the observer of the source output. 

Disorder is not enough of a measure for complexity,  as the throw of a dice or, for that matter, of a coin, conclusively shows \cite{hogg86}. We need a measure of storage and transmission of information. To measure the transmission of information, excess entropy $E$, with different names, has been used in a number of contexts \cite{grassberger86,crutchfield12}. Excess entropy is the mutual information between two halves of the text bi-infinite string \cite{grassberger86,feldman03}. 
\begin{equation}
\begin{array}{ll}
 E&=I(\ldots, s_{-1}: s_0, s_1, \ldots)\\\\
  &=\sum\limits_{n=1}^{\infty}\left(h_{\mu(n)}-h_\mu\right)\label{eq:ee}
 \end{array}
\end{equation}
where $h_\mu(n)=H[n]-H[n-1]$. The excess entropy is not a direct measure of information storage, but it is a measures of the amount of information stored on one halve useful to predict the other half.

Complexity measure now comes as the balance between information production, measured by $h_\mu$, and useful information correlation, measured by $E$. If a system is dominated by the overwhelming production of new information (large $h_\mu$), then there is little capacity for making use of past information to predict future behavior, $E$ will tend to zero. If, on the other hand, almost no new information is produced and therefore, $h_\mu \approx 0$, then the text is highly redundant (perhaps repetitive), and the dominant process will be transmitting information from earlier portions of the document to later portions ($E$ will be larger). Between both extremes, there is only a limited amount of excess entropy that a given entropy rate can accommodate. Complexity-entropy maps, or diagrams, readily shows the balance attained by any processing system between measures of disorder and useful storage, and have been used in many contexts \cite{grassberger86,feldman08,crutchfield12,sikai18}.  In this work, we will use the plot of $E$ vs. $h_\mu$ as the complexity-entropy diagram.

\subsection{Lempel Ziv estimation of entropy density and excess entropy}

Entropy rate and excess entropy have been defined for bi-infinite strings, which makes them impossible to calculate if the underlying structure (stochastic computational machinery) of the source is unknown. They have therefore to be empirically estimated. There is a vast literature about the ways to estimate entropy rate, and less for estimating excess entropy for a finite string of symbols, which we will not go into details here. Suffice to say that a factorization procedure as described by Lempel and Ziv \cite{lz76} was used as the basis for the estimation. The Lempel-Ziv factorization used in this study is related to, but must not be confused, with estimates based on compression rate as those done using public compression software. Our procedure is more robust and mathematically sounder for shorter string lengths than such estimates based on compression. 

There are different variants of Lempel-Ziv estimation of entropic measures. They are all based on the factorization of a symbol sequence following a set of rules. In this contribution, the factorization used is the one described in \cite{lz76}. A short description follows.

Let us denote $s(i,j)$ a subsequence that starts at position $i$ and ends at position $j$ then we define the $\pi$ operator as 
\begin{equation} 
s(i,j) \pi=s(i,j-1). \nonumber
\end{equation}

The string $s$ has an exhaustive factorization $F(s)$ 
\begin{equation}
F(s)=s(1,l_1)s(l_1+1,l_{2})\dots s(l_{m-1}+1,N), \nonumber
\end{equation}
if each factor $s(l_{k-1}+1,l_k)$ complies with
\begin{enumerate}
 \item $s(l_{k-1}+1,l_k)\pi\subset s(1,l_k)\pi^2$
 \item $s(l_{k-1}+1,l_k)\not\subset s(1,l_k)\pi$ except, eventually, for the last factor $s(l_{m-1}+1,N)$.
\end{enumerate}
This factorization is unique.

According to these rules, for example, the sequence $u=10010101110$ will have a factorization $F(s)=1.0.01.01011.10$, where each factor is delimited by a dot. The same factorization rules are followed for natural text, as for example in: 

F(s)=t.o. .s.h.e.r.l.oc.k. h.ol.m.es. she .i.s a.lw.ay.s t.he w.om.an

The Lempel-Ziv complexity $C_{LZ}(s)$ of the string $s$, is defined as the number of factors. In the binary example above, $C_{LZ}(s)$=5.

It has been shown \cite{ziv78} that the entropy density $h_\mu$ is given by 
\begin{equation}
  h_\mu=\limsup_{n\rightarrow\infty} \frac{C_{LZ}(s)}{n/\log{n}}.\label{eq:hlz}
\end{equation}
and can be estimated for long enough sequences \cite{lesne09,estevez13}.

For estimating the excess entropy an expression related to expression [\ref{eq:ee}] is used:
\begin{equation}
 E_{LZ}=\sum\limits_{M=1}^{M_{m}}(h_{LZ}(S_{(M)})-h_{LZ}(S)),\label{eq:elz}
\end{equation}
where $S_{(M)}$ is a surrogate sequence obtained by partitioning the string $S$ in non-overlapping blocks of length $M$, and performing a random shuffling of the blocks. The shuffling, for a given block of length $M$, destroys all correlations between symbols for lengths larger than $M$, while keeping the same symbol frequency. $M_{m}$ is chosen appropriately, given the sequence length, as to avoid fluctuations due to the sequence finite length $N$. We used $M_m=40$ and also $80$, with no significant change in the results. In spite that $E_{LZ}$ is not strictly equivalent to the excess entropy, as given by equation [\ref{eq:ee}], it is expected to behave in a similar manner\cite{melchert15}.

Calculations of both Lempel-Ziv complexity and $E_{LZ}$ were made by in-house software, already used in the study of cellular automata \cite{estevez15} which runs in $O(n \log n)$ time.

To emphasize that $h_\mu$ and $E$ are estimated from the Lempel-Ziv factorization, the symbols $h_{LZ}$ and $E_{LZ}$ are used instead. 

\section{Text sources}

As already mentioned, three English speaking authors were chosen: William Shakespeare, Arthur Conan Doyle, and Jacob Abbot\footnote{A grasp at each author work and style can be found at Wikipedia (www.wikipedia.org)}.  The source of all text was the public Gutenberg project at https://www.gutenberg.org. Abbott is mostly a writer of children and youngster books, with also known biographies and religious books. Conan Doyle is the father of Sherlock Holmes, but also a  writer of historical novels and some war books and chronicles. His works were mostly written for newspapers. His style is therefore direct with not much use of elaborate metaphors or involved language, as was directed towards the general public. Shakespeare, on the other hand, is considered by most critics the pinnacle of English literature and a master of language.

All texts used in the analysis are publicly available (See \ref{shakestbl}, \ref{doyletbl} and \ref{abbottbl} for a complete list of the texts used).  Each text was filtered before estimations of entropy rate and excess entropy was performed. Filtering consisted of changing all letters to lower case; all diacritics were removed leaving the base character. All numbers were removed. Also, all punctuation marks were changed to the space character except for ``?'' and ``!'' that were changed for periods before any scrambling of sentences was performed when appropriate, after which periods were also changed for spaces. The resulting alphabet consisted of 27 characters. All texts were cut to the same length of $78000$ characters. $41$ texts were used for Shakespeare and Doyle, while the number of Abbott's text was $45$. 

In the case of plays, like Shakespeare texts, additional filtering was performed. All text that was not intended to be spoken by the potential actors was removed, this included the description of scenarios, actions by the characters of the play, scene names, and the name of each character before their parliament.

\section{Results}

Three types of results will be discussed. First, the complexity-entropy diagram for the works of three authors will show how individual aspects of their literary work and style are captured by the maps. Then, several randomizing experiments over the original text will be performed. At the sentence level, sentences will be shuffled randomly losing all correlation in the sentence organization hierarchy and above. Sentence shuffling has been done before using other types of analysis \cite{pavlov01}. At the word level, a third data set will be derived from the original works by performing randomization by word. In the case of word randomization, all correlations at and above the word organization hierarchy are lost. At the letter level, a set of surrogate texts will be derived by randomly shuffling the characters in the original documents. Character randomization loses all correlations not derived exclusively from the frequency of use of each letter. The complexity-entropy map will again be the tool of choice to study all three surrogate texts. Finally, the information contained in the sentence, word and character organizations will be evaluated.

\begin{figure}
\centering
\includegraphics[width=.5\linewidth]{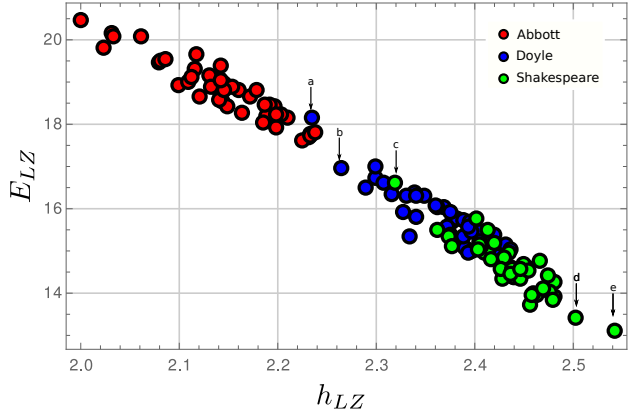}
\caption{Complexity-entropy map for the three studied authors. Points marked by arrows corresponds to (a) `` The coming of the fairies'', (b) ``The war in Southafrica'', (c) ``The merry wives of Windsor'', (d) ``The tempest'' and (e) ``The rape of Lucrece''.}
\label{fig:em}
\end{figure}

Our first result is shown in Figure \ref{fig:em}, where the complexity-entropy map for the texts of all three authors can be seen. The most relevant fact is that authors are segmented in different regions of the diagram. The maximum possible value for the entropy rate is $h_{max}=\log 27=4.75$ bit/character, corresponding to a random organization of all possible symbols in the alphabet. For all authors, the entropy rate is far from this limit, which is expected for any intelligible text. While Abbott shows the smaller $h_{LZ}$ values and the larger $E_{LZ}$ values, Shakespeare, at the other end, exhibits the largest and smaller entropy density and $E_{LZ}$, respectively. Abbott texts are highly redundant and less innovative regarding information creation, which corresponds to an author that is writing for young readers with a mostly limited vocabulary (See Table \ref{tbl:aths}), simple syntax structures and also simple content organization. Shakespeare, on the other hand, is less tied to simple grammar and content organization, with a more vast vocabulary. Abbott's book, with the smallest vocabulary, has $0.75$ fraction of the vocabulary of Shakespeare text with fewer different words. Doyle lies between the other two authors, with a center of mass more towards Shakespeare. Noticing that Doyle vocabulary is as large as that of Shakespeare (Table \ref{tbl:aths}), the explanation for the grouping difference has to be found at a level above the vocabulary used. 

\begin{table}[!ht]
{\centering
\caption{Author statistics}
\begin{tabular}{lrrrrrrrr}
Author      &  T   & $\langle V \rangle$ &  $\sigma(V)$ & $V_{mi}$ & $V_{ma}$   & $\langle S \rangle $ & $S_{mi}$ & $S_{ma}$   \\\midrule %
Abbott      &  45  &        2110         &     352      &    1427   & 2609        &    8.55              &    8.12   &   8.88   \\
Doyle       &  41  &        2769         &     208      &    2314   & 3192        &    8.96              &    8.63   &   9.20    \\
Shakes. &  41  &        2673         &     216      &    2195   & 3270        &    9.14              &    8.78   &   9.58    \\
\bottomrule
\end{tabular}\label{tbl:aths}
}
\begin{tabular}{ll}
$T$: &Total number of books.\\
$\langle V \rangle$: &Mean size of the vocabulary over all analyzed text.\\
$\sigma(V)$: &Standard deviation over the vocabulary.\\
$V_{mi}$: &Smallest vocabulary.\\
$V_{ma}$: &Largest vocabulary.\\
$\langle S \rangle$:&Mean value of the entropy over word use histogram.\\
$S_{mi}$:&Smallest entropy over word use histogram.\\
$S_{ma}$:&Largest entropy over word use histogram.
\end{tabular}
\end{table}

For each author, there are very few outliers, actually just some isolated points. For Shakespeare, the isolated value with higher $E_{LZ}$ and smaller $h_{LZ}$ corresponds to ``The merry wives of Windsor'' (point (c) in Fig. \ref{fig:em}), while the two isolated points with the largest $h_{LZ}$ are from ``The tempest'' (point (d)) and ``The rape of Lucrece'' (point (e)). There is a consensus that ``The tempest'', the last sole work of Shakespeare, is, stylistically speaking, a play different from all other works of this author with a more organized and strict neoclassic cannon. ``The rape of Lucrece'' is a sonnet. The entropy map can distinguish the peculiar character of these two texts. When looking into Doyle, also the isolated point with the largest $E_{LZ}$ (point (a)) is not a novel or a short story, but a pretended study made by the author to show the existence of fairies from some photographic evidence. The next isolated point (b) is from a war chronicle, highly structured and redundant. Again the complexity-entropy diagram can distinguish these two works from the bulk of its other short stories and novels.

\begin{figure}
\centering
\includegraphics[width=.5\linewidth]{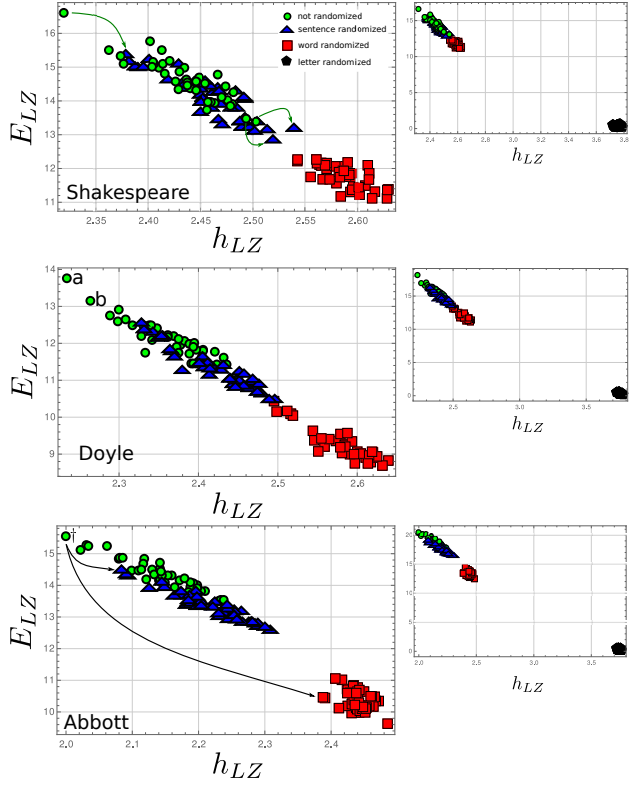}
\caption{Complexity-entropy map for the three studied authors. Arrows points where particular text ended up after randomization showing that randomization does not preserve the same order in the diagram.}
\label{fig:emauth}
\end{figure}

Next, texts were randomized by sentence, words, and letters. Figure \ref{fig:emauth} shows the complexity-entropy diagram for each author separately.   For all randomizations, sorting order of the original data is not kept for either $h_{LZ}$ and $E_{LZ}$.  

Word randomization destroys any organization hierarchy beyond word level. Entropy rate, in this case, measures pattern production from letters to words and will depend mostly on the vocabulary and the frequency on the use of words in each text. Abbott exhibits, for the bulk of its writing, the smallest $h_{LZ}$ in correspondence with the simplest vocabulary (Table \ref{tbl:aths}). Doyle and Shakespeare with similar vocabulary size show similar entropy rate for the central part of their work, yet six texts from Doyle are separated from the rest by its lower $h_{LZ}$ values. These points correspond to writings which are not novels or short stories: three war chronicles and three texts over spiritualism. These six texts show the smaller number of words among those of Doyle. The mean value of $h_{LZ}$ for the word randomized texts, for each author, follows the same order as the average entropy over the word use histogram (Table \ref{tbl:aths}), with Abbott having the smallest value. It should also be noticed that Abbott has the smallest spreading of the $h_{LZ}$ values, showing the more homogeneous character of his works concerning the use of vocabulary and words frequency. Regarding the behavior of $E_{LZ}$ for the word randomized texts, it essentially measures the random distribution of words along the text, which in turn is a function of the word frequency. 

To test if the analysis for word randomized text is meaningful from a linguistic point of view, we also performed some analysis in artificial words using as delimiters of a word, instead of space, the character ``e''. There is also a shift in the values of entropy density and $E_{LZ}$ (See \ref{fig:eword}), as should be expected, but results and text ordering in the complexity-entropy map changes, showing that the analysis is not independent on how a word is defined. This behavior can be indicative that the results for word-randomized text are useful from a linguistic analysis perspective.

The fact that the complexity-entropy map for the sentence randomized texts shows a shift towards higher values of $h_{LZ}$ and smaller values of $E_{LZ}$ for all three authors, is indicative of organizational structure at the levels above sentences. The most significant shift in Shakespeare text is found for ``The second part of King Henry the sixth'', where the relative decrease in $E_{LZ}$ reaches nearly $9\%$ and in $h_{\mu}$, $2.4\%$. Compare this values with Doyle, where ``The coming of the fairies'' has the largest relative shift with $5.2\%$ and $14.0 \%$ for  $h_{LZ}$ and $E_{LZ}$, respectively; or with Abbott where ``Rollo in London'' has an entropy density relative shift of $6.5 \%$, and ``Margaret of Anjou'' has the largest $E_{LZ}$ relative shift with $13.6 \%$. One can observe that in the case of Abbott, there is a significant gap in the complexity-entropy diagram between the points for the sentence and word randomized texts. 

We also studied if there was any dependence of $h_{LZ}$ and $E_{LZ}$ on sentence length. We found none for Shakespeare and Doyle, but for Abbott, there is a tendency for larger entropy density, and smaller $E_{LZ}$ the larger is the entropy over the sentence size. Again, such a trend could not be identified for Shakespeare and Doyle (See \ref{fig:ent}).

Randomization by character destroys all correlations above those determined by letter use frequency, correspondingly, there is a significant jump of $h_{LZ}$ towards higher values still below the maximum value of $4.75$, and of $E_{LZ}$ towards the neighborhood of zero. The entropy rate values are in the range of $3.6\sim3.7$  more in accordance with the estimates made in \cite{schurmann99}. The drop of $E_{LZ}$ to the almost zero measures the loss of correlations at all length scale.

\begin{figure}
\centering
\includegraphics[width=.2\linewidth]{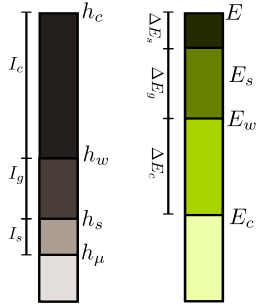}
\caption{Diagrams showing the different contribution to the entropy density (left) and excess entropy (right). $h_{LZ}$ and $E_{LZ}$ are the entropy rate and excess entropy of the original text, respectively, while $h_s$ ($h_w$) and $E_s$ ($E_w$) are the entropy rate and excess entropy for the sentence (word) randomized text. Finally, the entropy density for the character randomized text is denoted by $h_c$ and for the same text the excess entropy goes as $E_c$.}
\label{fig:diffd}
\end{figure}

In order to separate the various contributions to the entropy rate consider the difference
\begin{equation}
\begin{array}{l}
 I_s=h_s-h_{LZ}\\\\
 I_w=h_w-h_s\\\\
 I_c=h_c-h_w,
  \end{array}
\end{equation}
where $h_s$, $h_w$ and $h_c$ are the entropy rates for the sentence, word and character randomised texts, respectively. $I_s$ can be interpreted as the amount of information that goes into organising the sentences in the original text. Equivalently, $I_w$ measures the amount of information that goes into organising the words within the sentences of the original text, and $I_c$ measures the same magnitude but for organising letters to form words regardless of their order (Fig. \ref{fig:diffd}). 

Similarly, we define the excess entropy gain from organizing the sentences in the original text $\Delta E_s$, from organizing the words within the sentences of the original text $\Delta E_w$, and  from organizing the characters within the words of the original text $\Delta E_c$, as
\begin{equation}
\begin{array}{l}
 \Delta E_s=E_{LZ}-E_s\\\\
 \Delta E_w=E_s-E_w\\\\
 \Delta E_c=E_w-E_c,
  \end{array}
\end{equation}

\begin{figure}
\centering
\includegraphics[width=.5\linewidth]{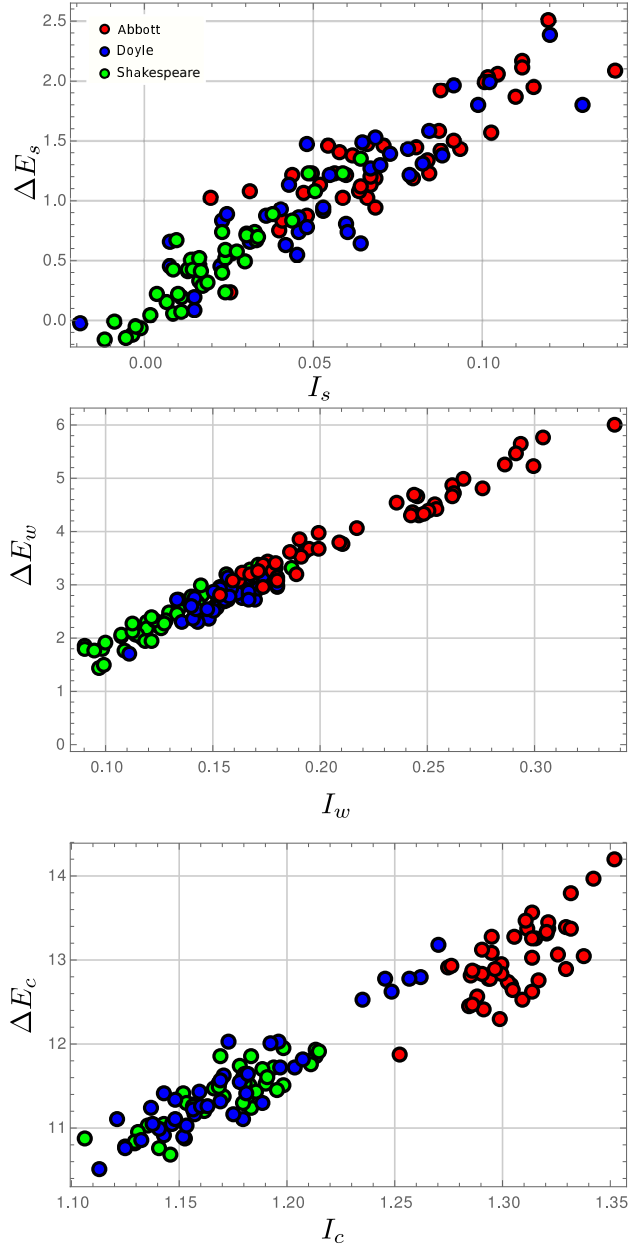}
\caption{Excess entropy gain vs Information gain for (above) sentence organization, (middle) word organization within sentences, (below) letter organization within words.}
\label{fig:emdiff}
\end{figure}

\begin{figure}[!ht]
\centering
\includegraphics[width=0.6\linewidth]{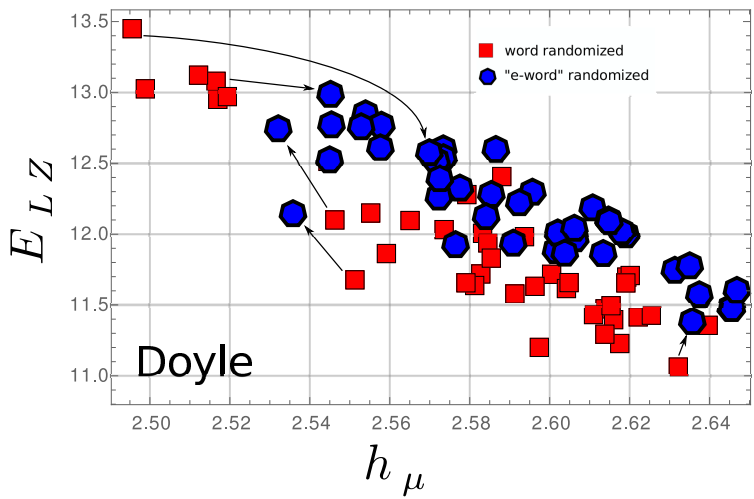}
\caption{Entropy-complexity map for word randomized text and ``e-word'' randomized text, showing that "structure" is different between them. ``e-words'' are artificial words using the character ``e'' as delimiter.}\label{fig:eword}
\end{figure}

\begin{figure}[!ht]
\centering
\includegraphics[width=0.8\linewidth]{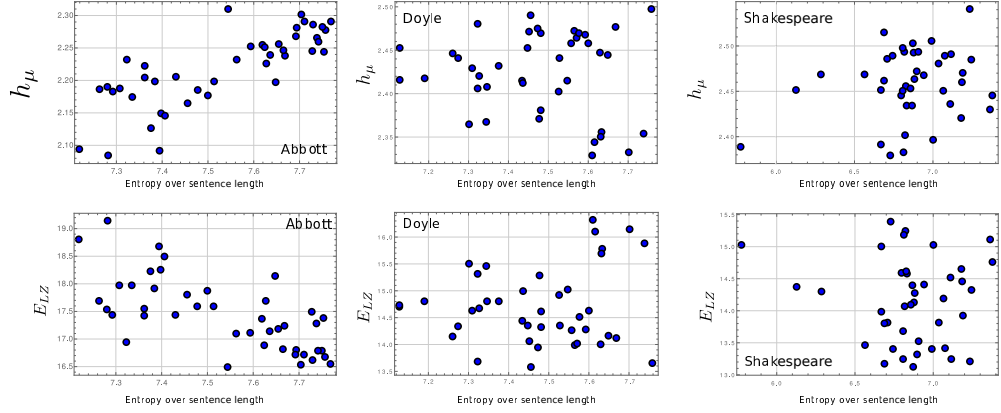}
\caption{Entropy density and excess entropy plotted against the entropy over the sentence length, for the sentence randomized texts of all three authors.}
\label{fig:ent}
\end{figure}

In figure \ref{fig:emdiff} one sees that the values of information gain $I_s$ can go from almost zero to around $0.14$ bit/character. Shakespeare values are the ones gathered around the smallest values of $I_s$, which also corresponds to the smallest values of excess entropy gain (center of mass $(I_s,\Delta E_s)=(0.0197,0.442)$), while Doyle is more spread over the region of values with center of mass $(I_s,\Delta E_s)=(0.0536,1.021)$ and then to the right, are Abbott texts with center of mass $(I_s,\Delta E_s)=(0.0728,1.364)$. Although Shakespeare appears more clearly segmented to the left, the values of figure \ref{fig:emdiff}(above) suggest that the pair $(I_s,\Delta E_s)$ are not enough to distinguish between authors. The center of mass for each author points to the fact that Shakespeare plays are less dependent on its correlation on the sentence organization, than Doyle and even so for Abbott. To which extent this can be linked to the rationale and style within each work is less evident, and must be done, if possible at all, with extreme care.  Let us say that $I_s$ shows, on average, that innovation for Shakespeare, as a process of pattern production, has more to do with what happens within sentences than in the relation between sentences, as compared to the other authors. For each text, the balance between innovation in sentence organization, as measured by $I_s$, and information correlation between sentences, as shown by $\Delta E_s$, can exhibit a range of variation, much in the sense of the varied nature of each author own works.

A similar analysis can be carried out for $(I_w,\Delta E_w)$. In this case segmentation between authors is more pronounced (Figure \ref{fig:emdiff} middle). The center of mass for each author corresponds to $(0.22,4.07)$ for Abbott, $(0.16, 2.79)$ for Doyle and $(0.12, 2.35)$ for Shakespeare. Abbott shows the largest spread of values, while Doyle the less. It must be noticed that for a given value of information gain, there is a larger spread in the values of the excess entropy it can accommodate for $\Delta E_s$ than for $\Delta E_w$. The values in figure \ref{fig:emdiff}(middle) are more cluttered around a straight line that fits the data than what can be seen for figure \ref{fig:emdiff}(above).  

Finally, $I_c$ and $\Delta E_c$ are the largest compared to the other information gain and excess entropy gain. It is not trivial that even in such case, Abbott separates from the other two authors, a fact that can be attributed to the word redundancy in Abbott's work as captured by the entropy over word use (See table \ref{tbl:aths}). The more extensive vocabulary in use for Shakespeare and Doyle means that once words are formed, larger entropy remains in these two authors compared to Abbott.

\section{Discussion}

A trivial behavior can have two flavors: a rule-based pattern production with no uncertainty, as in a periodic sequence, or a completely random process such as the toss of a coin. A useful text certainly must be between the two extremes.  A text dominated exclusively by information production, as measured by a maximum $h_{\mu}$ value, will look to the reader as a not very logically driven message, any context is lost from one section to another. A text dominated exclusively by the excess entropy will be as redundant as boring. After reading a finite portion of the text, the reader can skip the rest as conveying no new information that can not be derived unambiguously from what has been read already.

Predictability in a text can have different sources. Grammar, syntax, and semantics determine a set of rules that result in some amount of predictable behavior. So does the fact that a limited vocabulary is always in use. Also important is that in a rational and intelligent message a specific content in a given context is trying to be conveyed by the author. All these factors result in a redundant use of information. There is also the matter of the information flow in a text; one expect certain logic in the construction of a message, where references are made to earlier information written in the same document: there is a flow of information from previous parts in the text as context needed to understand later parts of the same message. Even if complying with all these constraints, there is the source of unpredictability coming from the known fact that there are a high number of different ways the same subject, content, or in general message, can be written. 

Even if predictability in natural language comes from constraints and the redundant use of information, this predictability could be built in different ways. When studying complexity-entropy diagrams in spin systems under local interaction dynamics, it is clear that for a given value of entropy density $h_\mu$, there is a range in possible values of excess entropy. This range can go even from a maximum value down to zero. The same can be seen in $4$-states Markov chains or cellular automata \cite{feldman08,crutchfield12}. From those comparisons, it comes as an interesting feature of the complexity-entropy diagrams shown here, that it seems that in all cases, for the original works and the randomized sentence texts, the value of $E_{LZ}$ shows a narrow spread of value for a given entropy density. Again this has not been the case in other systems as the mentioned examples show. One could infer that in written natural language the system stays near specific values of the excess entropy within a given unpredictability, which is also near a possible maximum value\footnote{One must be careful though, our data comes from known writers and we cannot exclude the possibility that including inferior English writing can broaden the range of allowed excess entropy values for a given entropy density.}. The question rises if maximization of excess entropy keeping a constant entropy density is a trend in natural written language. In figure \ref{fig:emauth}, one can see that for the word randomized text, there is an increase in the range of $E_{LZ}$ values for a given $h_{LZ}$, pointing to a loss of such optimization. The other interesting feature, related to the above comment is that, as figure \ref{fig:em} shows, all texts lie along the same straight line with slope $-13.87$ and intercept $48.70$. We analyzed books from over $100$ English speaking writers and the straight line held for all this diversity of sources. It could be asked if for different languages the same linear fit can be used and if so, the fitted parameters are universal or change from one language to another. 

Relevant to the above discussion is a similar analysis performed recently in art paintings \cite{sikai18}. Using a different complexity measure, the authors analyzed complexity-entropy maps of a large number of authors and paintings spanning over a millennium. Their complexity measure is used to assert repetitions of ordinal patterns in the image and consequently can also be related to redundancy. Similar to our result, the complexity exhibits a narrow spread of value for a given entropy density (See figure 2 in \cite{sikai18}) much in the same fashion as our result. It is remarkable that even if paintings are less constraint in the formal rules as written text, a similar diagram emerges. It may be asked if any intelligent message follows the same trend of accommodating a narrow range of complexity for given randomness.  The pair complexity-entropy was related to painting style, and therefore style in paintings seem to correlate with the balance between (dis)order and structure. A relevant question also for literary analysis. 

If we turn to information and excess entropy gain, the largest values happen for the letter organization within words (Figure \ref{fig:emdiff} middle). Words are the basic units of the language, so $I_c$ is measuring the entropy gain going from a letter ordering, which conveys none, or at least very little, information at all, to the first form of organization, from which an intelligent agent can extract some useful information. This information gain also comes as the result of establishing correlations between portions of the text which explains the large $\Delta E_c$ value. The successive gains at the word and sentence levels are decreasingly smaller in both information and excess entropy: organizing the words in sentences has a more significant effect in terms of information and excess entropy than organizing sentences into the final text. The later gives a quantitative measure to an idea that complies with our intuition.

\begin{figure}[!ht]
\centering
\includegraphics[width=0.6\linewidth]{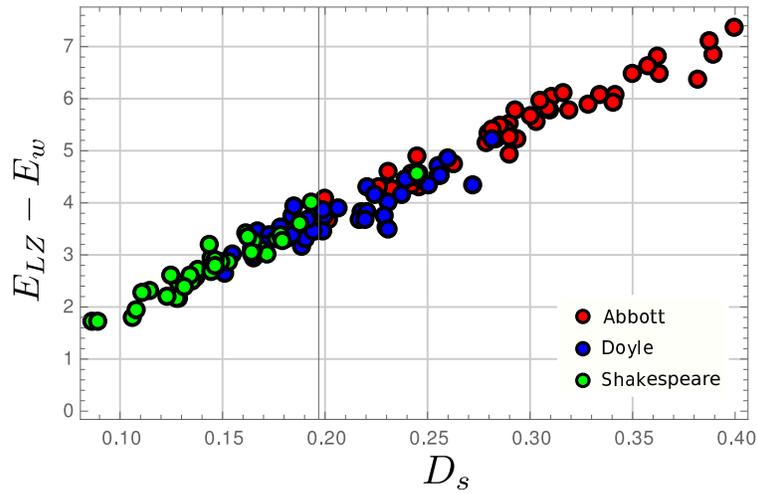}
\caption{ $(E_{LZ}-E_w)$ vs $D_s(=h_{w}-h_{LZ})$, showing a clear segmentation by authors which points to the non universality of $D_s$.}\label{fig:ds}
\end{figure}

Montemurro and Zanette \cite{montemurro15} has used $D_s=h_\mu-h_w$ to propose that this measure behaves as a universal constant value across different languages and even language families. As figure \ref{fig:emdiff} shows, the variations of $I_w$ are not random fluctuations resulting from a unique probability distribution for a given language, at least for English, instead,  $I_w$ seems to be biased towards different mean values for different authors. The same type of bias was found for $D_s$ (See \ref{fig:ds}). Given this result, the claim of Montemurro and Zanette has to be reevaluated in its meaning, as $D_s$ seems to be, at least, author dependent. 

\section*{Conclusions}
 
The interplay between all factors involved in natural languages, subjective and objective, makes written-text a highly complex data consequence of a highly complex source. 

The presented analysis shows that it is possible to, at least partially, assert the balance between innovation, as measured by entropy rate, and structure or context preservation, as measured by excess entropy. Within the scope of natural language, context preservation is associated with the rationality of the message conveyed by the data in the sense of the flow of information from one part of the text to another. Innovation can be related to the production of information.

This complexity organizes at different levels, starting from words, to sentences, to paragraphs and so on, up to the entire text. Is it possible to separate in written natural language, all possible formal and logical constraints from literary creativity? We believe that the answer is yes, at least to some extent, as the results presented here shows. Innovation as the production of new patterns, here followed by $h_{LZ}$, is a partial measure of such capacity. Even more, it is possible to follow innovation at different levels such as in words organization within sentences, followed by $I_w$, and at the level of sentence organization, followed by $I_s$. It is clear that for word randomized data the information that remains pertains mainly to the vocabulary used.

We have shown that, furthermore, by suitable manipulation of the data, the contribution from different levels of organization can be separated. One can notice that this separation was possible because separators for different structural units of information were known (spaces for words, periods for sentences). There are many long, complex data, where the knowledge of separators are known, and the analysis here performed can be generalized to such systems. Consider for example DNA as a string of a four-letter alphabet, where it is known that there is a hierarchy of information organization at different scales, starting from three letters codons to genes and beyond. A number of separators are known to exist in DNA at different scales. The tools developed could prove to be useful to identify the interplay of correlation and innovation in DNA and to measure the contribution to these two measures at different scales of organization.

It may be tempting to use the same measures as used here, namely excess entropy and entropy density, to study sources of short messages. It must be considered that for short sequences, such as in for example tweets, the error in the estimates of the measures may prove significant to make them meaningful\cite{lesne09}. Further research into short message sources is needed to have a better understanding of the behavior of the used measures with message length, and its comparison to other used methods \cite{amancio15}.

The results presented here also opens some question that will be the subject of further research. For example, if given a vocabulary and its frequencies are fixed, how meaningful are the texts that maximizes the excess entropy and minimizes entropy rate?

\section{Supporting information}

\label{shakestbl}
{\bf List of all text used from William Shakespeare.}
\begin{table}[H]
\centering
\bigskip
\begin{tabular}{ll}
 The two gentlemen of Verona                   & King Henry VI, First Part \\
 King Henry VI, Second Part                    & King Henry VI, Third Part \\
 King Richard III                              & The First Part of King Henry the Sixth \\
 The Rape of Lucrece                           & The Second part of King Henry the Six \\
 King Richard the Second                       & Loves Labours lost \\
 The life and death of King Richard the Second & A Midsummer Nights Dream \\
 The First Part of King Henry the Fourth       & Romeo and Juliet \\
 The life and death of King John               & The Merchant of Venice \\
 The Second Part of King Henry IV              & King Henry the Fifth \\
 Much Ado About Nothing                        & The Life of King Henry the Fifth \\
 The Taming of the Shrew                       & Julius Caesar \\
 As You Like It                                & The Merry Wives of Windsor \\
 The Tragedie of Hamlet                        & Twelfe Night, Or what you will \\
 The History of Troilus and Cressida           & Alls Well, that Ends Well \\
 The Tragedy of Othello, Moor of Venice        & Measure For Measure \\
 The Tragedie of Anthonie, and Cleopatra       & The Tragedie of King Lear \\
 Pericles                                      & The Life of Timon of Athens \\
 The Tragedy of Coriolanus                     & Sonnets \\
 Cymbeline                                     & The Winters Tale \\
 The Life of Henry VIII                        & The tempest \\
 The third part of King Henry the Six          &
\end{tabular}
\end{table}

\label{doyletbl}
{\bf List of all text used from Arthur C. Doyle.}
\begin{table}[H]
\centering
\bigskip
\begin{tabular}{ll}
  A Duet, with an Occasional Chorus                  & A Study in Scarlet \\
  Beyond the City                                    & Danger and Other Stories \\
  Micah Clarke                                       & Round the Fire Stories \\
  Sir Nigel                                          & Tales of Terror and Mystery \\
  The Adventures of Gerard                           & The Adventures of Sherlock Holmes \\
  The Captain of the Polestar, and Other Tales       & The Coming of the Fairies \\
  The Dealings of Captain Sharkey, and               &\\
  Other Tales of Pirates                             & The Doings of Raffles Haw \\
  The Exploits of Brigadier Gerard                   & The Firm of Girdlestone \\
  The German War                                     & The Great Boer War \\
  The Great Keinplatz Experiment and Other Tales     &\\ 
  of Twilight and the Unseen                         & The Great Shadow and Other\\
                                                     & Napoleonic Tales\\
  The Green Flag, and Other Stories of War and Sport & Through the Magic Door \\
  The Gully of Bluemansdyke, and Other stories       & The Hound of the Baskervilles \\
  The Last Galley; Impressions and Tales             & The Last of the Legions and\\
                                                     & Other Tales of Long Ago \\
  The Lost World                                     & The Man from Archangel, and\\
                                                     & Other Tales of Adventure \\
  The Memoirs of Sherlock Holmes                     & The Mystery of Cloomber \\
  The New Revelation                                 & The Poison Belt \\
  The Refugees                                       & The Return of Sherlock Holmes \\
  The Sign of the Four                               & The Tragedy of the Korosko \\
  The Valley of Fear                                 & The Vital Message \\
  The Wanderings of a Spiritualist                   & The War in South Africa, Its Cause \\
                                                     & and Conduct \\
  The White Company &   
\end{tabular}
\end{table}

\label{abbottbl}
{\bf List of all text used from Jacob Abbott.}
\begin{table}[H]
\centering
\bigskip
\begin{tabular}{ll}
 Alexander the Great Makers of History          &  Bruno \\
 Caleb in the Country                           & Charles I \\
 Cleopatra                                      & Cousin Lucys Conversations \\
 Cyrus the Great                                & Darius the Great \\
 Genghis Khan, Makers of History Series         & Hannibal \\
 History of Cleopatra, Queen of Egypt           & History of Julius Caesar \\
 History of King Charles the Second of England  & Jonas on a Farm in Winter \\
 King Alfred of England                         & Marco Pauls Voyages and\\
                                                & Travels\\
 Margaret of Anjou                              & Mary Erskine \\
 Mary Queen of Scots                            & Nero \\
 Peter the Great                                & Pyrrhus \\
 Queen Elizabeth                                & Richard I \\
 Richard II                                     & Richard III \\
 Rollo at Play Or, Safe Amusements              & Rollo at Work \\
 Rollo in Geneva                                & Rollo in Holland \\
 Rollo in London                                & Rollo in Naples \\
 Rollo in Paris                                 & Rollo in Rome \\
 Rollo in Scotland                              & Rollo in Switzerland \\
 Rollo on the Atlantic                          & Rollo on the Rhine \\
 Rollos Experiments                             & Rollos Museum \\
 Rollos Philosophy                              & Romulus \\
 The Teacher                                    & William the Conqueror \\
 Xerxes                                         & Vermont
\end{tabular}
\end{table}

\section{Acknowledgments}

This work has been partially supported by the Cuban Ministry of Higher Education and the University of Havana under a research project grant LEMPELCOMPLEX.

The authors would like to thank  the referees, who carefully read the manuscript and made valuable suggestions.


\begin{thebibliography}{10}

\bibitem{nowak01}
M.~A. Nowak and N.~L. Komarova.
\newblock Towards an evolutionary theory of language.
\newblock {\em Trends in Cognitive Sci.}, 5:288--295, 2001.

\bibitem{nowak02}
M.~A. Nowak, N.~L. Komarova, and P.~Niyogi.
\newblock Computational and evolutionary aspects of language.
\newblock {\em Nature}, 417:611--617, 2002.

\bibitem{montemurro15}
M.~A. Montemurro and D.~H. Zanette.
\newblock Universal entropy of word ordering across lingusitic families.
\newblock {\em Plos One}, 6:e19875, 2011.

\bibitem{chomski65}
N.~Chomski.
\newblock {\em Aspects of the theory of syntax.}
\newblock MIT Press, Cambridge, Mass., 1965.

\bibitem{hawkins83}
J.~A. Hawkins.
\newblock {\em Word order universals.}
\newblock Academic Press, New York, 1983.

\bibitem{hawkins04}
J.~A. Hawkins.
\newblock {\em Efficiency and complexity in grammars.}
\newblock Oxford University Press, Oxford, 2004.

\bibitem{debowski11}
L.~Debowski.
\newblock Excess entropy in natural language: Present state and perspective.
\newblock {\em Chaos}, 21:037105, 2011.

\bibitem{altmann11}
E.~G. Altmann, G.~Cristadoro, and M.~D. Esposti.
\newblock On the origin of long-range correlations in texts.
\newblock {\em Proc.Natl. Acad. Sci. USA},
  doi.org/10.1073/pnas.1117723109:310--324, 2011.

\bibitem{lacalle06}
E.~Alvarez-Lacalle, B.~Dorow, J.~P. Eckmann, and E.~Moses.
\newblock Hierarchical structures induce long-range dynamical correlation in
  written texts.
\newblock {\em Proc.Natl. Acad. Sci. USA}, 103:7956--7961, 2006.

\bibitem{montemurro10}
M.~A. Montemurro and D.~Zanette.
\newblock Towards the quantification of the semantic information encoded in
  written language.
\newblock {\em Adv. Complex Syst.}, 13:135--153, 2010.

\bibitem{shannon51}
C.~Shannon.
\newblock Prediction and entropy of printed english.
\newblock {\em Bell Syst. Tech. J.}, 30:50--64, 1951.

\bibitem{amancio13}
D.~R. Amancio, E.~G. Atmann, D.~Rybski, O.~N. Oliveira, and L.~da~F.~Costa.
\newblock Probing the statistical properties of unknown texts: Application to
  the voynich manuscripts.
\newblock {\em PLOS}, 8:e67310, 2013.

\bibitem{amancio15b}
D.~R. Amancio.
\newblock A complex network approach to stylometry.
\newblock {\em PLOS}, 10:e0136076, 2015.

\bibitem{ebeling95}
W.~Ebeling and T.~Poschel.
\newblock Entropy, transinformation and word distribution of
  information–carrying sequences.
\newblock {\em Int. J. Bifurcation and Chaos}, 5:51--61, 1995.

\bibitem{shannon48}
C.~Shannon.
\newblock A mathematical theory of communication.
\newblock {\em Bell Syst. Tech. J.}, 30:379--423, 1951.

\bibitem{schurmann99}
T.~Schurmann and P.~Grassberg.
\newblock Entropy estimation of symbol sequence.
\newblock {\em Chaos}, 6:414--427, 1999.

\bibitem{schurmann07}
T.~Schurmann and P.~Grassberg.
\newblock The predictability of leters in written english.
\newblock {\em Fractals}, 4:doi.org/10.1142/S0218348X96000029, 1996.

\bibitem{zipf65}
G.~K. Zipf.
\newblock {\em The Psycho-Biology of Language: An Introduction to Dynamic
  Philology.}
\newblock MIT Press, Cambridge, Mass., 1965.

\bibitem{mandelbrot54}
B.~Mandelbrot.
\newblock Structure formelle des texte et communication.
\newblock {\em Word}, 10:1--27, 1954.

\bibitem{herdan64}
G.~Herdan.
\newblock {\em Quantitative Linguistic}.
\newblock Butterworths, 1964.

\bibitem{heaps78}
H.~S. Heaps.
\newblock {\em Information retrieval-Computational and theoretical aspects}.
\newblock Academic Press, 1978.

\bibitem{crutchfield89}
J.~P. Crutchfield and K.~Young.
\newblock Inferring statistical complexity.
\newblock {\em Phys. Rev. Lett.}, 63:105--108, 1989.

\bibitem{crutchfield12}
J.~P. Crutchfield.
\newblock Between order and chaos.
\newblock {\em Nature}, 8:17--24, 2012.

\bibitem{jimenez16}
M.A. Jimenez-Montano, H.F. Coronel-Brizio, A.R. Hernandez-Montoya, and
  A.~Ramos-Fernandez.
\newblock Codon information value and codon transition-probability
  distributions in short-term evolution.
\newblock {\em Physics A}, 454:117--128, 2016.

\bibitem{cover06}
T.~M. Cover and J.~A. Thomas.
\newblock {\em Elements of information theory. Second edition}.
\newblock Wiley Interscience, New Jersey, 2006.

\bibitem{lz77}
J.~Ziv and A.~Lempel.
\newblock A universal algorithm for sequential data compression.
\newblock {\em IEEE Trans. on Info. Theory.}, 23:337--343, 1977.

\bibitem{hogg86}
B.~Huberman and T.~Hogg.
\newblock Complexity and adaptation.
\newblock {\em Physica D}, 22:376, 1986.

\bibitem{grassberger86}
P.~Grassberger.
\newblock Towards a quantitative theory of self-generated complexity.
\newblock {\em Int. J. Theo. Phys.}, 25:907--938, 1986.

\bibitem{feldman03}
J.~Crutchfield and D.~P. Feldman.
\newblock Regularities unseen, randomness observed: Levels of entropy
  convergence.
\newblock {\em Chaos}, 13:25--54, 2003.

\bibitem{feldman08}
D.~P. Feldman, C.~S. McTeque, and J.~P. Crutchfield.
\newblock The organization of intrinsic computation: complexity-entropy
  diagrams and the diversity of natural information processing.
\newblock {\em Chaos}, 18:043106--043121, 2008.

\bibitem{sikai18}
H.~Y.~D. Sigaki, M.~Perc, and H.~V. Ribeiro.
\newblock History of art painting through the lens of entropy and complexity.
\newblock {\em PNAS}, 115:E8585--E8594, 2018.

\bibitem{lz76}
A.~Lempel and J.~Ziv.
\newblock On the complexity of finite sequences.
\newblock {\em IEEE Trans. Inf. Th.}, IT-22:75--81, 1976.

\bibitem{ziv78}
J.~Ziv.
\newblock Coding theorems for individual sequences.
\newblock {\em IEEE Trans. Inf. Th.}, IT-24:405--412, 1978.

\bibitem{lesne09}
A.~Lesne, J.~l.~Blanc, and L.~Pezard.
\newblock Entropy estimation of very short symbolic sequences.
\newblock {\em Phys. Rev. E}, 79:046208--046218, 2009.

\bibitem{estevez13}
E.~Estevez-Rams, R.~Lora Serrano, B.~Aragon Fernandez, and I.~Brito Reyes.
\newblock On the non-randomness of maximum lempel ziv complexity sequences of
  finite size.
\newblock {\em Chaos}, 23:023118--023124, 2013.

\bibitem{melchert15}
O.~Melchert and A.~K. Hartmann.
\newblock Analysis of the phase transition in the two-dimensional ising
  ferromagnet using a lempel-ziv string-parsing scheme and black-box
  data-compression utilities.
\newblock {\em Phys. Rev. E}, 91:023306--023317, 2015.

\bibitem{estevez15}
E.~Estevez-Rams, R.~Lora-Serrano, C.~A.~J. Nunes, and B.~Arag\'on-Fern\'andez.
\newblock {L}empel-{Z}iv complexity analysis of one dimensional cellular
  automata.
\newblock {\em Chaos}, 25:123106--123116, 2015.

\bibitem{Note1}
A grasp at each author work and style can be found at Wikipedia
  (www.wikipedia.org).

\bibitem{pavlov01}
A.~N. Pavlov, W.~Ebeling, L.~Molgedey, A.~R. Ziganshin, and V.~S. Anishchenko.
\newblock Scaling features of texts, images and time series.
\newblock {\em Physica A}, 300:310--324, 2001.

\bibitem{Note2}
One must be careful though, our data comes from known writers and we cannot
  exclude the possibility that including inferior English writing can broaden
  the range of allowed excess entropy values for a given entropy density.

\bibitem{amancio15}
D.~R. Amancio.
\newblock Probing the topological properties of complex networks modeling short
  written texts.
\newblock {\em PLOS}, 10:e0118394, 2015.

\end{thebibliography}

\end{document}